\documentclass[10pt,twocolumn,letterpaper]{article}

\usepackage[pagenumbers]{wacv}      

\definecolor{wacvblue}{rgb}{0.21,0.49,0.74}
\usepackage[pagebackref,breaklinks,colorlinks,allcolors=wacvblue]{hyperref}

\def\wacvPaperID{1982} 
\def\confName{WACV}
\def\confYear{2026}

\title{Stroke Modeling Enables Vectorized Character Generation with Large Vectorized Glyph Model}

\author{Xinyue Zhang$^{\ddagger}$ \\
Tongji University \\
{\tt\small elainexinyuezhang@gmail.com}
\and
Haolong Li$^{\ddagger}$ \\
Tongji University \\
{\tt\small Furlongli322@gmail.com}
\and
Jiawei Ma \\
Tongji University \\
{\tt\small jiavve1.10@gmail.com}
\and
Chen Ye$^{*}$ \\
Tongji University \\
{\tt\small yechen@tongji.edu.cn}
}

\begin{document}
\maketitle
\begin{abstract}
Vectorized glyphs are widely used in poster design, network animation, art display, and various other fields due to their scalability and flexibility. In typography, they are often seen as special sequences composed of ordered strokes. This concept extends to the token sequence prediction abilities of large language models (LLMs), enabling vectorized character generation through stroke modeling. In this paper, we propose a novel Large Vectorized Glyph Model (LVGM) designed to generate vectorized Chinese glyphs by predicting the next stroke. 
Initially, we encode strokes into discrete latent variables called stroke embeddings. Subsequently, we train our LVGM via fine-tuning DeepSeek LLM by predicting the next stroke embedding.
With limited strokes given, it can generate complete characters, semantically elegant words, and even unseen verses in vectorized form. Moreover, we release a new large-scale Chinese SVG dataset containing $907,267$ samples based on strokes for dynamically vectorized glyph generation. Experimental results show that our model has scaling behaviors on data scales. Our generated vectorized glyphs have been validated by experts and relevant individuals.
\end{abstract}
    
\section{Introduction}
\label{sec:intro}

\begin{figure}[t]
  \includegraphics[width=\columnwidth]{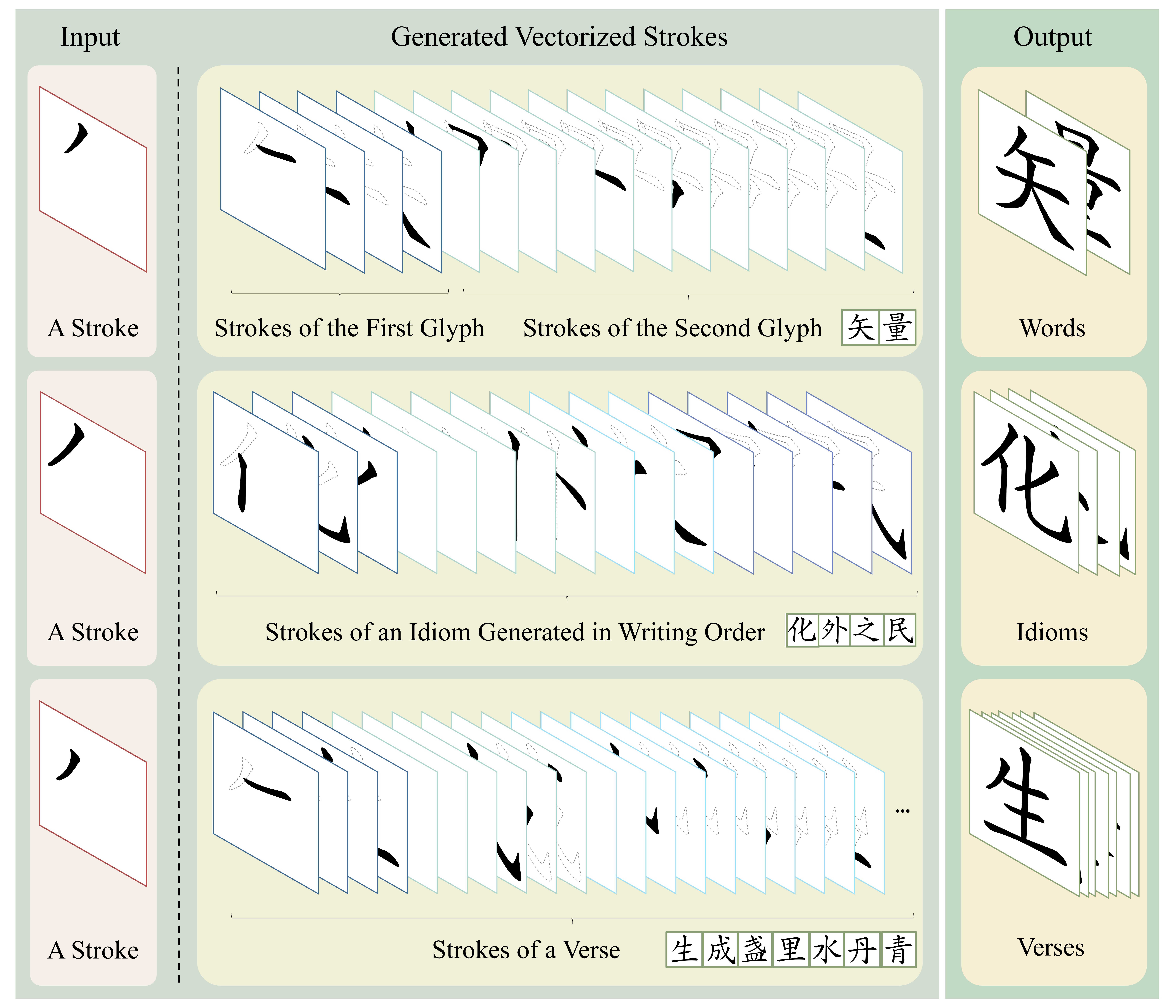}
  \caption{Examples of different vectorized cases generated by Large Vectorized Glyph Model (LVGM), only using a stroke.}
  \label{fig:example3}
\end{figure}

Characters serve as a cornerstone of human expression and knowledge. Whether engraved on ancient stone monuments \cite{li2024towards,li2021generative} or displayed on digital screens, characters preserve the essence of cultures, ideologies, and personal narratives. Characters nowadays are usually represented in two forms: rasterized glyphs and vectorized glyphs (In typography, a glyph is "the specific shape, design, or representation of a character"\footnote{Glyph: https://en.wikipedia.org/wiki/Glyph}). Rasterized glyphs, made up of pixels arranged regularly, may distort and blur when scaled. Vectorized glyphs, which define characters using mathematical primitives (2D points or curves) in a resolution-independent fashion \cite{differentiable}, have been widely used for poster design, artistic displays, and digital animations due to their scalability.

Many prior researches\cite{aoki2022svg,xing2024svgdreamer,zhang2023beyond,rodriguez2023starvector} have explored vectorized generation. Im2Vec\cite{Im2vec} and DeepSVG\cite{Deepsvg} generate high quality vector graphics under raster image supervision and hierarchical generative network based on a two-stage seq-to-seq model. These methods view simple patterns as basic units and create intricate vectorized images by organizing units in a structured sequence. Similarly, glyphs can be viewed as a special sequence composed of ordered strokes—the basic unit of glyphs-just as we write characters stroke by stroke in daily lives.  Each stroke, meticulously crafted, contributes not only to the visual form of a glyph but also carries semantic weight and historical significance. However, existing methods\cite{svg-vector-font-generation,DeepVecFont,DeepVecFont-v2,Deepsvg,nagata2023contour,VecFontSDF}, generate glyphs in segments and then stitch them together, which leads to a lack of coherence in overall glyph structure. To address this limitation, some explorations, aiming to simulate the natural stroke-by-stroke writing process and recreate the organic flow and rhythm present in authentic handwritten characters, have become a necessity.

Large Language Models (LLMs) have shown excellent performance across a variety of natural language tasks. GPT4\cite{gpt-4}, LLaMA\cite{llama}, Gemini\cite{gemini}, Minerva\cite{minerva}, Mistral\cite{mistral} and DeepSeek\cite{deepseek-llm,deepseek-coder,deepseek-vl} have emerged as dominant models in popular benchmarks\cite{li-exploringmath}. Many field-specific LLMs have exhibited remarkable performance within their respective areas of expertise. ChatCounselor\cite{ChatCounselor} is a customized LLaMA-7B model fine-tuned with counseling domain instruction data for mental health support. DISC-LawLLM\cite{DISC-LawLLM} is an intelligent legal system to provide a wide range of legal services. They have shown remarkable performance in respective field. Can we also create a large language model for generating vectorized glyphs? Can this model simulate the real writing process by predicting the next stroke? Furthermore, can such a model generate more semantically elegant words or even poems in vectorized form given a few strokes?

In this paper, we introduce a novel and pioneering model, named Large Vectorized Glyph Model (LVGM), which is the first to utilize a GPT-like architecture for vectorized character generation, marking a significant advancement in this field. LVGM, a two-stage stroke-learning model, generates vectorized glyphs by predicting the next stroke. Initially, we collect a new dataset of vectorized Chinese characters, containing a large number of authoritative and representative words, idioms, and verses, which contributes a lot to vectorized glyph generation work and character design. We then standardize all SVG commands into a sequence of cubic Bezier curves represented by coordinates, called stroke representation for our two-stage training process. In Stage \uppercase\expandafter{1}, we encode the stroke representation into a discrete latent variable known as stroke embedding, achieving high-fidelity reconstruction. The Stage \uppercase\expandafter{2} involves fine-tuning a series of LLMs based on DeepSeek-Coder-1.3B\cite{deepseek-coder} to predict the next stroke embedding, culminating in the creation of our LVGM. Moreover, our model demonstrates scalability and notable performance gains with increasing dataset sizes. Remarkably, LVGM can generate complete Chinese glyphs in vectorized form from just a few initial strokes, encompassing not only visually appealing characters but also semantically rich words and verses absent from our training data, as shown in \ref{fig:example3}. Expert validation ensures the validity and quality of results generated by LVGM. We will make our dataset and model publicly available.

To sum up, our major contributions can be concluded as:
\begin{itemize}
    \item We propose a novel model called Large Vectorized Glyph Model (LVGM), the first GPT-like model of vectorized character generation to our knowledge. With just a few strokes as input, LVGM can generate complete Chinese glyphs, semantically coherent words, and even entire poems in vectorized form.
    \item We introduce a new large-scale Chinese SVG dataset containing 907,267 samples based on strokes. Each glyph is represented as a closed SVG path, which are stored in stroke order.
    \item Experiments indicate that increasing the amount of high-quality data leads to performance enhancements in identifiability, aesthetics, literary quality of our generated vectorized glyphs.
\end{itemize}
\section{Related work}

\subsection{Vectorized Character Generation}

In numerous scenarios, glyphs are stored in vector formats, offering scalability and occupying significantly less storage space than pixmaps, prompting extensive efforts toward vectorized generation.
SVG-VAE\cite{SVG-VAE} first attempts to generate SVG fonts using a sequential generation model, while Carlier et al.\cite{Deepsvg} proposed DeepSVG, a novel hierarchical generative network based on transformers. Reddy et al.\cite{Im2vec} introduced Im2Vec, a model capable of generating complex vector graphics with different topological structures requiring only indirect supervision from raster images, while Wang et al.'s\cite{DeepVecFont} DeepVecFont employs a dual-modal learning approach, utilizing LSTM\cite{lstm} to process vector graphics and leveraging a differentiable rasterizer to improve the visual quality.
The aforementioned methods concentrate on generating vectorized Chinese characters at the character level,  which leads to a lack of coherence in the overall glyph structure. Our approach centers on the strokes directly, incrementally building entire characters by predicting each subsequent stroke in sequence.

\subsection{Large Language Models}

\begin{figure*}
    \centering
    \includegraphics[width=\linewidth]{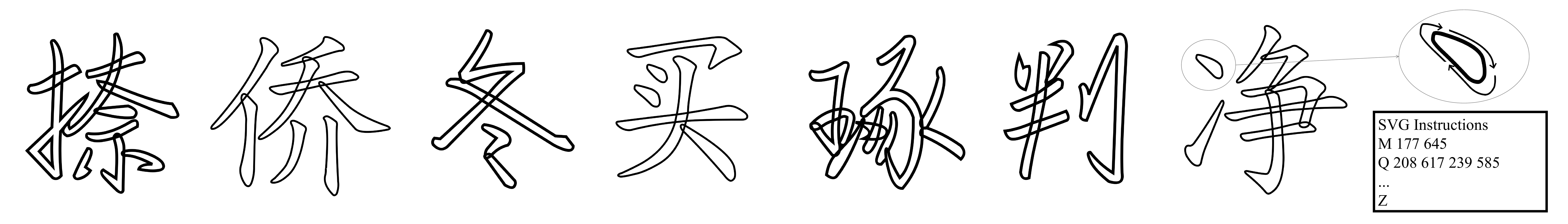}
    \caption{Samples from the SVG-Strokes dataset in vector graphics format. Characters are in multiple caligraphy styles and each stroke is annotated by a single closed curve. In our dataset, all SVG glyphs are outlined with parameterized drawing instructions, each stroke segmented into complete outlines, forming multiple SVG instructions for each contour.}
    \label{fig:datasample}
\end{figure*}
\begin{figure*}
\centering
\includegraphics[width=\textwidth]{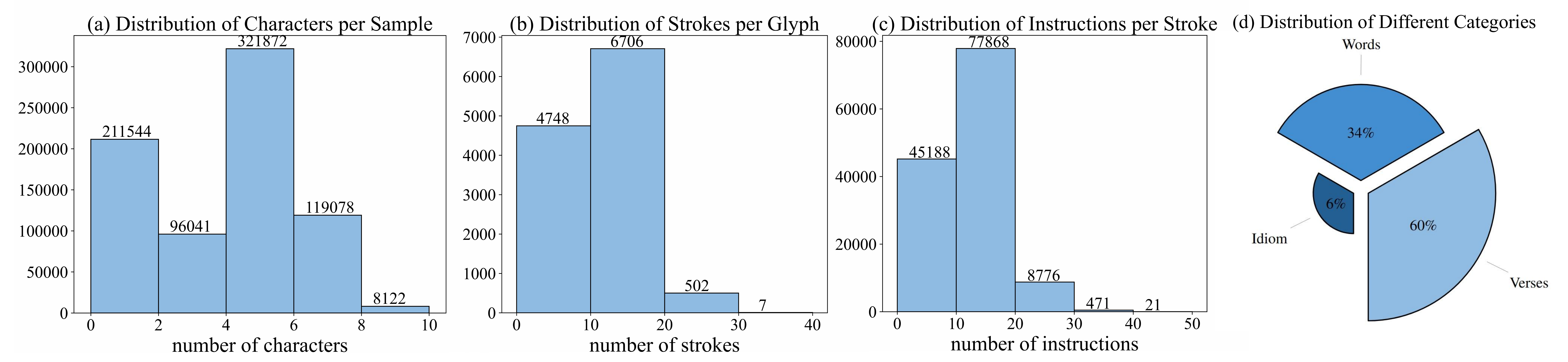}
\caption{Statistical distribution at different levels. (a) denotes the distribution of the number of characters at the sample level; (b) shows the distribution of the number of strokes at the character level; (c) displays the distribution of the number of instructions at the stroke level; (d) illustrates the distribution of different categories in our dataset.}
\label{fig:statistics}
\end{figure*}

In recent years, the domain of large language models (LLMs) has witnessed a remarkable evolution. This advancement has been fueled by the augmentation of training datasets and the escalation in model complexity\cite{llm-survey,scaling-law}. Pioneering models like BERT\cite{bert}, GPT-2\cite{gpt-2}, and T5\cite{t5} have played pivotal roles in the initial stages of large language model development, significantly propelling progress in the field\cite{minigpt4}. Subsequent adherence to scaling laws has led to the introduction of models with augmented parameter counts, exemplified by GPT-3\cite{gpt3}, PaLM\cite{palm}, LLaMA\cite{llama}, Gemini\cite{gemini}, DeepSeek\cite{deepseek-coder,deepseek-llm,deepseek-vl}, and GPT-4\cite{gpt-4}, among others. These models have showcased exceptional proficiency in natural language comprehension and the resolution of intricate tasks, thereby revolutionizing various language-related benchmarks. This success has also led to the development of specialized models for specific fields, such as Galactica\cite{galactica} for science, ChatCounselor\cite{ChatCounselor} for mental health support, and DISC-LawLLM\cite{DISC-LawLLM} for law. Building upon this trend, we were inspired to embark on the development of a new large language model centered on stroke-based approaches for the generation of vectorized glyphs. This effort demonstrates the broad potential of large language models beyond traditional language applications and into specialized domains.
\section{Dataset}

\subsection{SVG-Strokes Dataset}
Existing vector font datasets either only contain single style of calligraphy or consider different characters simply as vector graphics.
These datasets therefore lack detailed dissection of Chinese characters at the individual stroke level and specific stroke order information, which is significant to Chinese character generation  and computer aided character design.
Thus, we introduce a new dataset, called SVG-Strokes.
It's composed of Chinese characters in semi-cursive script and regular script formed by closed curves created through segmentation of individual strokes and sequential arrangement according to stroke order.
We collect the origin SVG data from \textit{makemeahanzi}\footnote{https://github.com/skishore/makemeahanzi} and \textit{FZSJ-XIAOSXS}\footnote{https://www.foundertype.com/index.php/FontInfo/index/id/5296}.
Manual annotations are applied in stroke segmentation and order setting.
Our dataset consists of $744,810$ samples in regular script, and $162,457$ semi-cursive ones, which are useful for dynamically vectorized glyph generation, shown in \ref{fig:datasample}.

\subsection{Data Format}
Different from rasterized graphics, vector graphics use fundamental drawing primitives to represent visual elements.
While Scalable Vector Graphics (SVG)\cite{eisenberg2014svg} is a widely adopted format for two-dimensional vector graphics, particularly known for its capability in vectorized character rendering, we employ this XML-based methodology to structure and organize our dataset's data structures.
In our dataset, every glyph is represented in strokes, with each stroke individually segmented into a complete outline. Each contour comprises multiple SVG instructions, as shown in \ref{fig:datasample}. Moreover, the SVG instructions used are listed in supplementary.

\subsection{Data Annotation}
Most calligraphy styles other than regular script exhibit significant challenges in stroke segmentation due to merging and connectedness issues. To address this, we design specific stroke segmentation rules for labeling our dataset:
\begin{itemize}
    \item If the start of one stroke does not align with the end of another stroke, they should be segmented separately.
    \item If the ending part of the preceding stroke is not characterized by a transition from thick to thin, these two strokes should be connected.
    \item If two strokes are connected (i.e., the start of one stroke aligns with the end of another), but the preceding stroke exhibits a transition from thick to thin, they should be segmented separately.
\end{itemize}

\begin{table}
  \centering
  \begin{tabular}{@{}|c|c|c|c|@{}}
    \toprule 
    \textbf{Category} & \textbf{Glyph} & \textbf{Total} & \textbf{Source}\\
    \midrule 
    Words & Regular & 246,042 & Xinhua Dictionary\\
    & Semi & 89,032 & Xinhua Dictionary\\
    \midrule
    Idioms & Regular & 29,797 & Xinhua Dictionary\\
    & Semi & 7,268 & Xinhua Dictionary\\
    \midrule
    Verses & Regular & 468,971 & Tang Poems\\
    & Semi & 66,157 & Tang Poems\\
    \bottomrule 
  \end{tabular}
  \label{tab:data-cate}
  \caption{The statistics of our dataset. 'Words' are normal vocabularies; 'Idioms' are phrases or expressions that have a figurative meaning that is different from the literal meanings of the individual words; 'Verses' refer to lines of text in a poem.}
\end{table}

\begin{figure*}[t]
\centering
\includegraphics[width=\linewidth]{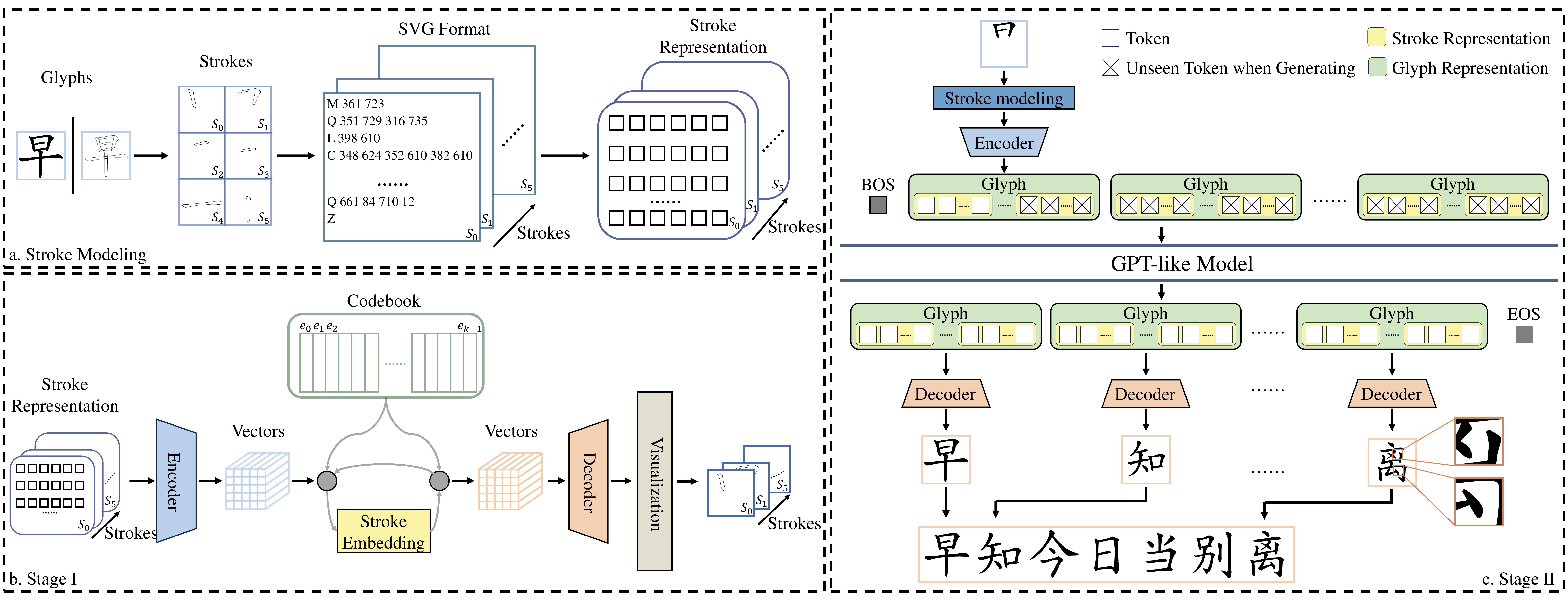}
\caption{Overview of our proposed method. (a) In our stroke modeling, every stroke constituting the glyph in SVG format, which is composed of a series of SVG instructions, is converted to a stroke representation. (b) We encode the stroke representation into the stroke embedding (discrete latent variable) and train a decoder for reconstruction in Stage \uppercase\expandafter{1}. (c) In Stage \uppercase\expandafter{2}, we fine-tune a GPT-like model to predict the next stroke embedding and generate the semantically elegant verse using the encoder and decoder trained in Stage \uppercase\expandafter{1}.}
\label{fig:pipeline}
\end{figure*}

\subsection{Data Statistics}
We compiled a comprehensive dataset of $907,267$ data extracted from reputable, authoritative, well-known, and representative sources, covering words, idioms, and verses with different lengths of characters. Detailed information is presented in 
\ref{fig:statistics}. 

We conduct a statistical analysis of the distribution of our proposed dataset.  As shown in \ref{fig:statistics}, the analysis revealed that the character lengths across various genres exhibit a relatively uniform distribution spanning from 1 to 10 characters. Furthermore, the maximum stroke count observed for a single Chinese glyph in regular script is 34 (for semi-curisve script is 24), and the highest number of SVG drawing instructions employed for a single stroke in regular script was determined to be 49 (for semi-cursive script is 79).
\section{Method}

\subsection{Stroke Modeling}
Existing studies on vectorized character generation exhibit a lack of coherence in the overall glyph structure, often generating segments independently and merging them without regard for stroke sequencing. We believe that generating glyphs based on stroke order, mirroring the process of writing, enhances aesthetic quality. This approach aligns with human cognition and emphasizes the importance of representing each stroke distinctly.  Given individuals' ingrained writing practices, characters can be conceptualized as a special sequence comprising ordered strokes. Therefore, we propose a simple but effective approach for stroke modeling.

We define a glyph $G$ to be a ordered sequence of $N_g$ strokes $S_i$.
\begin{equation}
    G=\{S_1, S_2, ...,S_{N_g}\}
\end{equation}
In standard SVG format, a stroke $S_i$ is composed of $N_s$ drawing instructions $I_i$ which are parameterized Bézier curves. 
\begin{equation}
    S_i=\{I_1, I_2, ..., I_{N_s}\}
\end{equation}

To simplify our training, we standardize all instructions $I_i$ to cubic Bézier curves $B_C(t)$ and then transform per curve into $6$ parameters $(x_{P_0}, y_{P_0}, x_{P_1}, y_{P_1}, x_{P_2}, y_{P_2})$, called $Para_i$. It's worth noting that $P_0$ of $I_i$ is the end point of the previous instruction $I_{i-1}$, which equals to its $P_3$. The $P_0$ of $I_0$ is from 'M' type instruction which always presents at each stroke's beginning. 
\begin{equation}
    \begin{aligned}
    B_C(t) = &(1-t)^3P_0 + 3t(1-t)^2P_1\\ &+ 3t^2(1-t)P_2 + t^3P_3,\ t\in[0, 1]
    \end{aligned}
    \label{eq:BC}
\end{equation}
\begin{equation}
    I_i \to Para_{i}(x_{P_0}, y_{P_0}, x_{P_1}, y_{P_1}, x_{P_2}, y_{P_2})
\end{equation}

Therefore, given a glyph $G$ is redefined by:
\begin{equation}
    G=\{Rep_{1}, Rep_2, ...,Rep_{N_g}\}
\end{equation}
\begin{equation}
    Rep_{i}=\{Para_{1}, Para_{2}, ...,Para_{N_s}\}
\end{equation}

where stroke representation $Rep_{i}$ is composed of $N_s - 1$ parameters groups $Para_i$. 

\begin{table*}[t]
  \centering
    \resizebox{\textwidth}{!}{
  \begin{tabular}{@{}|c|l|c|c|c|c|@{}}
    \toprule
     \textbf{Participants} & \textbf{Evaluation} & \textbf{Fine-tuned on 10K} & \textbf{Fine-tuned on 25K} & \textbf{Fine-tuned on 50K} & \textbf{Fine-tuned on 100K} \\
    \midrule
    Experts & Identifiability$\uparrow$ & 3.41 & 3.87 & 4.67 & \textbf{4.96} \\
    & Aesthetics$\uparrow$              & 3.44 & 3.22 & 4.13 & \textbf{4.33}\\
     & Literature Quality$\uparrow$     & 2.88 & 3.53 & 4.17 & \textbf{4.43} \\
    & Final Score$\uparrow$             & 3.26 & 3.57 & 4.47 & \textbf{4.67}\\
    \midrule
    Students & Identifiability$\uparrow$  & 3.57 & 3.92 & 4.74 & \textbf{5.00}\\
    & Aesthetics$\uparrow$                         & 3.63 & 3.47 & 4.38 & \textbf{4.57}\\
     & Literature Quality$\uparrow$                & 3.07 & 3.56 & 4.30 & \textbf{4.40} \\
    & Final Score$\uparrow$                        & 3.44 & 3.68 & 4.50 & \textbf{4.69}\\
    \midrule 
    All & Weighted Final Score$\uparrow$                   & 3.31 & 3.60 & 4.48 & \textbf{4.68}\\
    \bottomrule
  \end{tabular}
  }
  \caption{Scores on different metrics of the model fine-tuned with different training set sizes (10K / 25K / 50K / 100K samples). Experts and graduate students evaluate generated images separately and the scores are averaged and calculated. The best results are \textbf{highlighted}.}
  \label{tab:human-estimation}
\end{table*}

\begin{figure*}
    \centering
    \includegraphics[width=\textwidth]{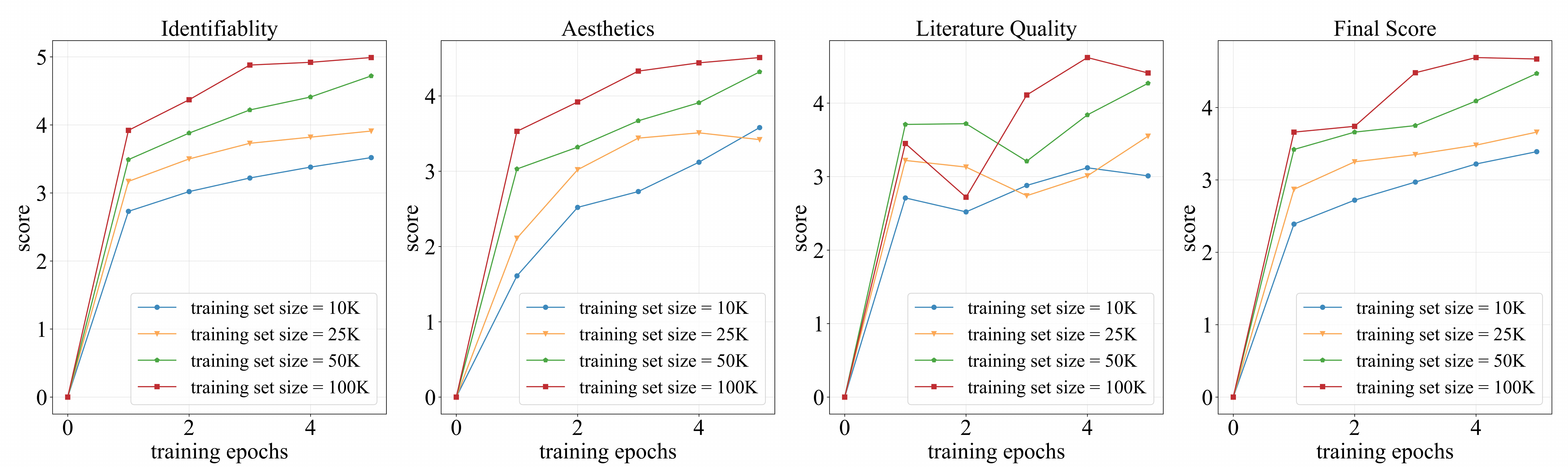}
    \caption{The score from three perspectives: identifiability, aesthetics, and literature quality, and the final score for different training set sizes (10K / 25K / 50K / 100K samples).}
    \label{fig:scores}
\end{figure*}

\begin{table}
  \centering
  \resizebox{\columnwidth}{!}{
  \begin{tabular}{@{}|c|cccc|@{}}
    \toprule 
    \textbf{Method} & \textbf{MSE$\downarrow$} & \textbf{PSNR$\uparrow$} & \textbf{SSIM$\uparrow$} & \textbf{LPIPS$\downarrow$}\\
    \midrule 
    DeepSVG & 0.1141 & 6.7540 & 0.4355 & 0.7363\\
    Im2Vec & 0.1032 & 9.6320 & 0.7423 & 0.5343\\
    DeepVecFont-v1 & 0.0643 & 12.3774 & 0.9103 & 0.1315\\
    DeepVecFont-v2 & 0.0518 & 14.9932 & 0.9246 & 0.1108\\
    \midrule
    LVGM & \textbf{0.0197} & \textbf{17.1259} & \textbf{0.9432} & \textbf{0.0394}\\
    \bottomrule 
  \end{tabular}
  }
  \label{tab:contrast}
  \caption{Comparison of graphic quality for different vectorized character generation methods, which shows our model's high generation quality.}
\end{table}

\begin{figure}
  \centering
    \includegraphics[width=\columnwidth]{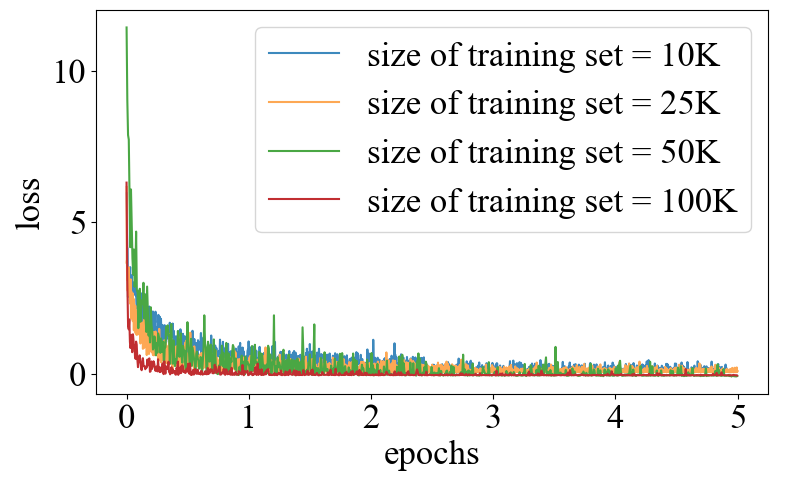}
   \caption{The training loss for different training set sizes (10K / 25K / 50K / 100K samples).}
   \label{fig:loss-epoch}
\end{figure}

\subsection{Stage \MakeUppercase{1}: Quantized vectors of learned stroke embeddings}
In stage \MakeUppercase{1}, our goal is to compress the stroke representation $Rep_{i}$ into a discrete variable and reconstruct the SVG instructions in a certain level of quality. Inspired by VQVAE\cite{van2017neural} and LVM\cite{bai2024sequential}, we use an encoder to compress our stroke modeling metric and regard the index of the nearest vector in the codebook as the compression results, which fits our goal. We employ the nearest neighbor search method to calculate the index of the $8$ fixed vectors in the codebook which is the closest to the $8$ vectors obtained by convolving each stroke. This process allows us to obtain the stroke embedding from the origin stroke representation.

We add ReLU\cite{glorot2011deep} layers between convolution layers and change the decoder structure into fractionally-strided convolutions, and then two residual blocks followed by two fractionally-strided conventions with a ReLU between them.

\subsection{Stage \MakeUppercase{2}: Strokes generation by order of writing}

In stage \MakeUppercase{ 2}, we organize the stroke embeddings in sequence for model training by writing orders and character orders, enabling the model to predict the next stroke embeddings which can represent strokes.
Constrained by experimental resources, we trained our model by fine-tuning DeepSeek-Coder-1.3B\cite{deepseek-coder}, which is a lightweight pretrained LLM with relatively good performance. We stack encoded stroke embeddings in stroke order and separate them by characters with a special token to form our stroke sequence for training. To control the overall context length, we systematically applied right-padding to input sequences. 

Our model is based on the implementation of TRL full stack library\cite{vonwerra2022trl} and uses FlashAttention v2\cite{dao2023flashattention} to speed up computations involved in the attention mechanism. 
\section{Experiment}
In this section, we evaluate our trained model, present scaling abilities, glyph completion ability, and multiple character generation ability.

\subsection{Settings}
In our experiments, all inputs and outputs data are SVG files. The encoder is normal residual blocks with hidden dim 1024, and the sampler of categorical variables to index codebook with $K=30000$ categories with embedding $dim=16$. More setting details are in appendix due to page limitations.
\subsubsection{Dataset}
In our experiments, all the input and output SVG data in regular script are limited to $64\times6$.
We set $N_g=34$, $N_s=64$ for stroke embedding considering data distribution.
The total number of Chinese characters is up to 9574. 
Locations of all SVG instructions are normalized to [-1, 1] during the training process for stability.
To ensure a balanced distribution of various types of data, we performed random sampling on each type of data in all experiments. This ensured that the training data consisted of 20\% words, 30\% idioms, and 50\% verses. 

\subsubsection{Framework}
The encoder to extract features from input data used in our method is normal residual blocks with hidden dim 1024 (residual hidden dims 256), and the sampling of categorical variables to index our codebook with $K=30000$ categories with embedding $dim=16$. 
To preserve the structural integrity of our data (formatted as $64\times 6$ matrices post-stroke modeling) and mitigate the risk of over-convolution along specific dimensions within Convolutional Neural Networks (CNNs), our preprocessing strategy involves reshaping the input data into $6\times 8\times 8$ matrices prior to feeding them into the encoder module. 
For the decoder, we use fractionally-strided convolutions and Relu to reconstruct images. 
With our encoder, one stroke after stroke modeling can be compressed to $8$ stroke embeddings, facilitating efficient data representation and compression.

\subsubsection{Implement Details}
We use two NVIDIA RTX A6000/A800 GPUs for all training and evaluating processes. An Adam with $lr=1e-4$ is used as the optimizer of the encoder and decoder. We trained our model with batch size 128 for 3000 iterations.
Limiting each data strokes sum under 100, the maximum total number of dataset mixed up with Chinese words, idioms and verses is 100K samples. 
We fine-tuned Deepseek-Coder-1.3B with the maximum allowable sum of strokes per individual sample as $N_c=100$ and the maximum text length parameter to $N_l=820$, label smoothing factor $0.001$, and batch size $4$ for $5$ iterations.

\subsection{Evaluation}
\label{subsec:evaluation}

Our method is highly innovative in simultaneously generating vectorized glyphs and text content. To measure complex character features, we designed the evaluation from three perspectives on identifiability, aesthetics, and literature quality, assigning scores ranging from 0 to 5 for each criterion. We invited 50 experts and 150 graduate students in relevant fields to participate in the evaluation, and the results of all assessments were aggregated. Indentifiability, aesthetics, linterature quality and score definition are explained in appendix.

\textbf{Identifiablity.}
It measures each glyph by the completeness of the strokes of the glyph, considering missing or incorrect strokes, whether the entire character is reasonable and carries a real meaning, and whether the glyph is easily recognizable as a character.

\textbf{Aesthetics.}
Focus on the smoothness of vector stroke edges, the presence of abnormal points and spikes, the correct relative positioning of each vector stroke, and the overall appearance quality of the glyph.

\textbf{Literature quality.}
In multi-word generation, the coherence of context is crucial. The presence of meaningless words can affect the overall effectiveness of the text. It is important to consider whether the generated content as a whole carries a certain level of literary quality.

We define formulas \ref{eq:score} to calculate our quality score for our model. We assign weights to different perspectives for a accurate final score $Score_{Final}$:
\begin{equation}
    Score_{Final} = 0.4\cdot Score_{Ide}+0.3\cdot Score_{Aes}+0.3\cdot Score_{Lit}
    \label{eq:score}
\end{equation}

We invited 50 experts and 150 graduate students in relevant fields to participate in the evaluation, and the results of all assessments were aggregated. All images were randomized. Participants were presented with images one by one and allowed to observe the images for 3 seconds before scoring within 5 seconds after observation \cite{zhang2018unreasonable}. We assign weights to the averaged scores given by experts $Score_{exp}$ and students $Score_{gs}$ to calculate the total score $Score$:

\begin{equation}
    Score = 0.7\cdot Score_{exp} + 0.3\cdot Score_{gs}
    \label{eq:final-score}
\end{equation}

\subsection{Scaling}
We investigate the scaling behavior of our model in terms of the training loss and the generation performance. 

We randomly sample 10K, 25K, 50K, and 100K samples from our collected dataset to explore the data scaling effects on the fine-tuned LVGM. The score on identifiability, aesthetics, Literature quality, and the final score of experts and graduate students are shown in \ref{tab:human-estimation}. When the model is fine-tuned with 100K samples, it achieves the highest scores with a final score of 4.68, 4.67 and 4.69 of experts and graduate students, respectively. Results show that our model gain better performance on the larger size training set.

Moreover, we show the training curves of the model fine-tuned on the dataset of random 10K, 25K, 50K, and 100K samples in \ref{fig:loss-epoch}. A faster decaying rate is clearly observed in the training loss when increasing the training data size, aligning with the marked rise in all scores measured by experts and graduate students in \ref{fig:scores}. These two experiments demonstrate our model's scaling behavior.

\subsection{Generation of Missing Strokes}

\begin{figure*}
\centering
\includegraphics[width=\textwidth]{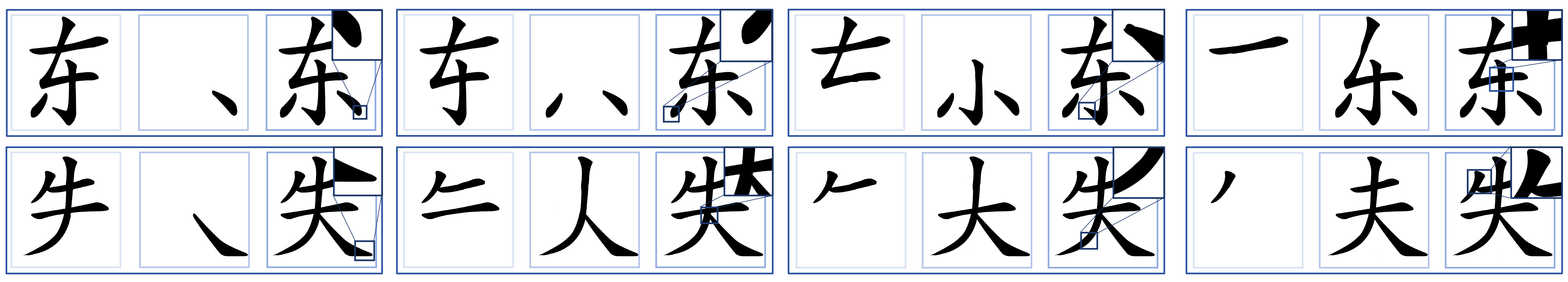}
\caption{Examples of the generation of missing strokes. The left frame is input data, the middle frame is the remaining output strokes that the model predicted, and the right frame is the generation results of our LVGM.}
\label{fig:missing-stroke}
\end{figure*}

\begin{figure*}
\centering
\includegraphics[width=\textwidth]{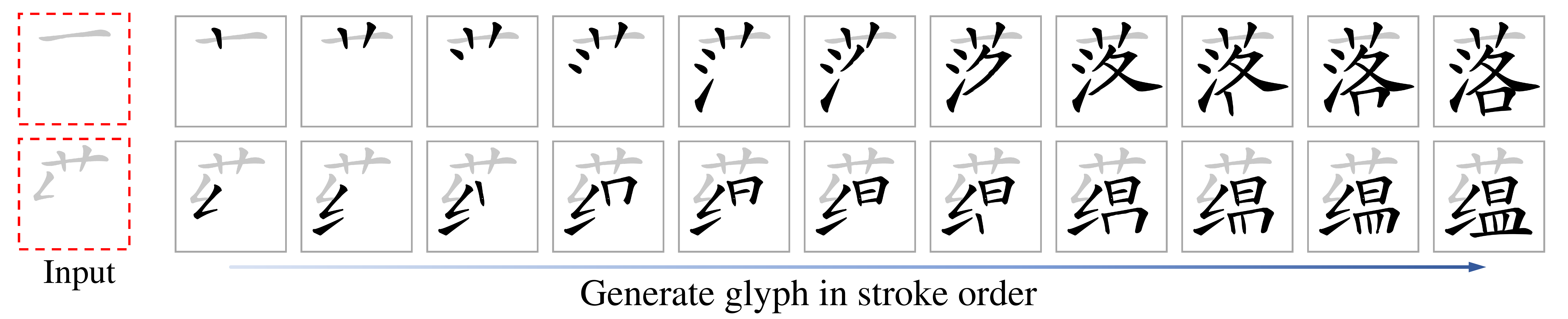}
\caption{Generating glyphs stroke by stroke. Gray strokes are inputs given to our model and black strokes are generated in stroke orders with high quality. Each line from left to right shows the generating order.}
\label{fig:inorder}
\end{figure*}

We investigate the ability of our LVGM to generate a glyph by providing limited-ordered strokes, and observing our predictions for the subsequent strokes. In 
\ref{fig:missing-stroke} and 
\ref{fig:inorder}, our model can predict the next strokes and complete the character only depending on the few strokes given. Even with a decreasing number of stroke hints, our model can still predict the remaining strokes and effectively generate correctly identifiable glyphs. It shows our models' strong ability to predict the next stroke and our high-quality generation of vectorized glyphs.

\subsection{Generation of multiple Chinese characters}

\begin{figure*}
\centering
\includegraphics[width=\textwidth]{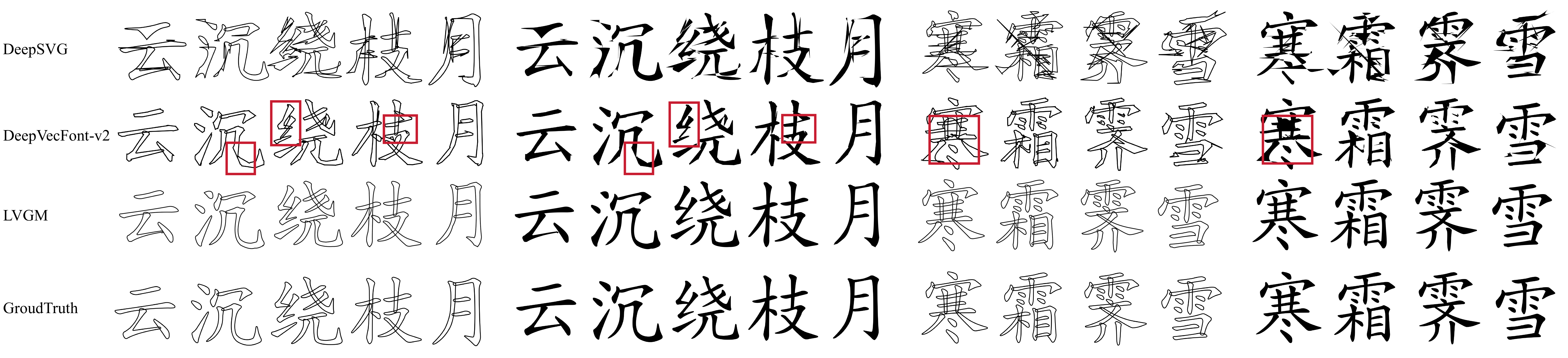}
\caption{Comparison of samples for character generation. Considering other models are not able to generate meaningful content, we give the text content generated by our LVGM and compare the vectorized outputs quality.}
\label{fig:contrast}
\end{figure*}

\begin{figure*}
\centering
\includegraphics[width=\textwidth]{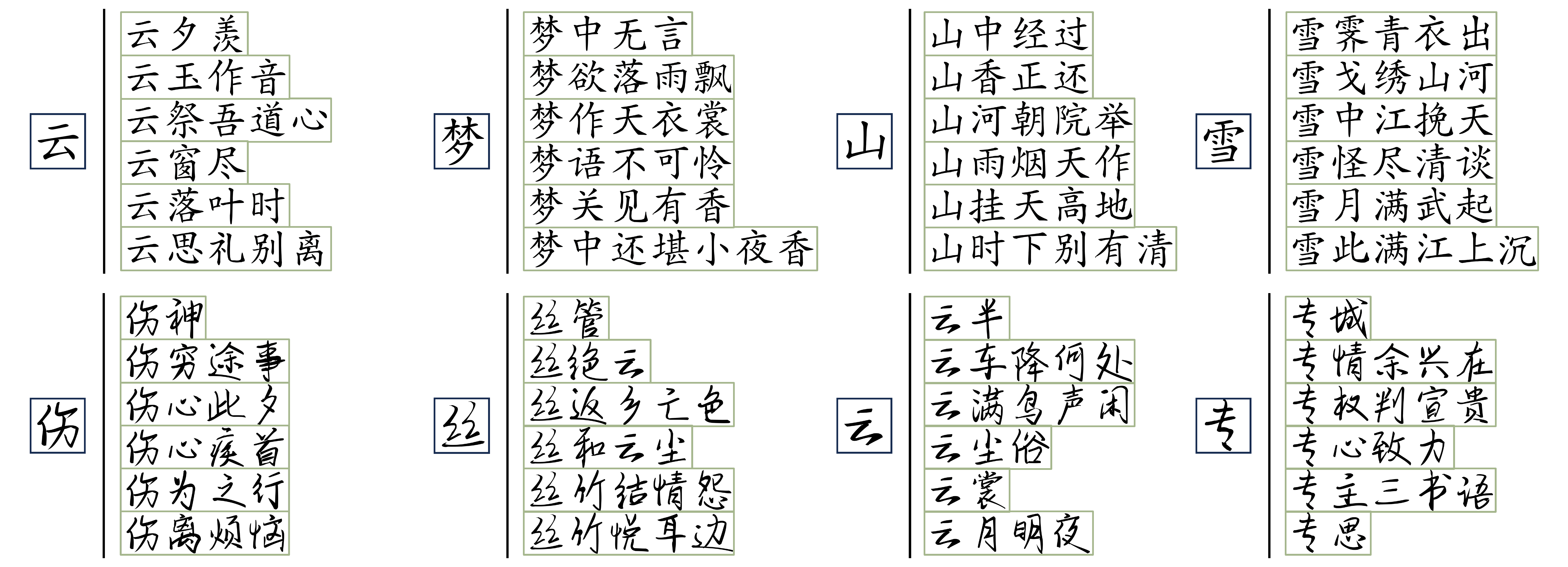}
\caption{Examples of the generation of unseen vectorized cases. Results show that our model excels in generating novel words and verses of literary and aesthetic value}
\label{fig:example2}
\end{figure*}

To expand the potential creativity of LVGM, we generate words and verses given a few stroke hints. As shown in \ref{fig:example2}, our model demonstrates remarkable creativity in generating previously unseen contents of literary and aesthetic appeal. These creations, hitherto unseen, evoke a sense of wonder and admiration among observers, highlighting the model's profound ability to transcend conventional boundaries in linguistic creativity. Given the content generated by our LVGM, we compare the generation results with other vector models in \ref{fig:contrast} and \ref{tab:contrast}, which demonstrates that LVGM has better vector quality.
\section{Discussion}
In this study, we have explored training Large Vectorized Glyph Model (LVGM) based on predicting the next stroke. While our model excels at generating remarkable vectorized results, it still exhibits shortcomings in following aspects: (1) Due to high compression rate in the first stage, the smoothness and completeness of the generated strokes may be lacking; (2) Given our limited training resources, we did not fine-tune pre-trained models of different sizes or explore larger datasets in our experiments. In the near future, we are going to extend our method: (1) Scale up both the model and the dataset and experiment with RL methods such as PPO and DPO; (2) Except for Chinese characters, the generation of glyphs in different languages and styles will be considered; (3) Multimodal data will be included. Notably, we adhere to strict ethical guidelines and principles in this research,  and our process and outcomes are free from intellectual property and ethical legal disputes.
\section{Conclusion}
Large language models (LLMs) are intrinsically zero-shot and multi-task learners, enabling exceptional performance in specialized domains and even vectorized character generation. In this paper, we introduce a novel Large Vectorized Glyph Model (LVGM), the first GPT-like model to generate vectorized glyphs based on stroke modeling and a new large-scale Chinese SVG dataset. Our LVGM can sequentially generate complete Chinese characters, even unseen words and verses in vectorized form stroke by stroke, starting from just a few initial strokes. Our data scaling experiment demonstrates that by increasing the amount of training data, the generation performance improves accordingly. 
We anticipate that the proposed LVGM will establish a foundational framework for future research in vectorized character generation and encourage exploration from a broader perspective in the field.

{
    \small
    \bibliographystyle{ieeenat_fullname}
    \bibliography{main}
}

\end{document}